\documentclass[10pt,conference,twocolumn,final,compsoc]{IEEEtran}

\usepackage[utf8]{inputenc}
\usepackage{enumitem}
\usepackage{amsmath}   
\usepackage{breqn} 
\usepackage{booktabs}  
\usepackage{multirow}  
\usepackage{siunitx}
\usepackage{subfig}
\usepackage{graphicx}
\usepackage[breaklinks=true]{hyperref}  
\usepackage[switch]{lineno}

\newsavebox{\measurebox}

\begin{document}

\title{Indirect Identification of Psychosocial Risks from Natural Language}

\author{
    \IEEEauthorblockN{Kristen C. Allen}
    \IEEEauthorblockA{Carnegie Mellon University\\
    \texttt{kcallen@cmu.edu}}
    \and
    \IEEEauthorblockN{Alex Davis}
    \IEEEauthorblockA{Carnegie Mellon University\\
    \texttt{alexdavis@cmu.edu}}
    \and
    \IEEEauthorblockN{Tamar Krishnamurti}
    \IEEEauthorblockA{University of Pittsburgh\\
    \texttt{tamark@pitt.edu}}
    } 

\maketitle

\begin{abstract}
During the perinatal period, psychosocial health risks, including depression and intimate partner violence, are associated with serious adverse health outcomes for parents and children. To appropriately intervene, healthcare professionals must first identify those at risk, yet stigma often prevents people from directly disclosing the information needed to prompt an assessment. We examine indirect methods of eliciting and analyzing information that could indicate psychosocial risks. Short diary entries by peripartum women exhibit thematic patterns, extracted by topic modeling, and emotional perspective, drawn from dictionary-informed sentiment features. Using these features, we use regularized regression to predict screening measures of depression and psychological aggression by an intimate partner. Journal text entries quantified through topic models and sentiment features show promise for depression prediction, with performance almost as good as closed-form questions. Text-based features were less useful for prediction of intimate partner violence, but moderately indirect multiple-choice questioning allowed for detection without explicit disclosure. Both methods may serve as an initial or complementary screening approach to detecting stigmatized risks. 
\end{abstract}


\section{Introduction}
Psychosocial risks experienced around pregnancy, including depression and intimate partner violence, can have meaningful consequences for the health and well-being of experiencers and their children. Depression significantly increases a woman's risk for adverse pregnancy outcomes, including low birth weight and intrauterine growth restriction \cite{grote2010meta}. A prospective study by Dayan \textit{et al.} found a significant correlation between preterm delivery and major depression measured using the Edinburgh Postnatal Depression Scale (EPDS) \cite{cox1987detection,Dayan2006}. The Women's Experience with Battering scale (WEB) \cite{Smith1999}, capturing the day-to-day lived experience of women subjected to intimate partner violence (IPV), has been linked to adverse pregnancy outcomes including preterm birth \cite{Campbell_2002}. Beyond acute health problems, a mother's psychosocial risks can affect interactions with her infant, as well as that infant's development \cite{grace2003effect}.

Early identification of psychosocial health risks allows for more effective intervention. Postpartum home visits by health professionals, peer-based postpartum telephone support, and interpersonal psychotherapy have shown reductions in postpartum depression \cite{howard2014non}. Simple screening programs can reduce the risk of depression, with or without additional treatment components \cite{o2016primary}. Primary care physicians are equipped to implement meaningful interventions in cases of IPV \cite{bair2014primary}. 

To implement interventions that address depression and intimate partner violence during the perinatal period, clinicians must first detect the problem. Yet stigma and social norms may prevent women from presenting clinicians the information needed to detect and intervene on psychosocial risks \cite{kingston2015barriers, overstreet2013intimate}. Pregnant women may have an elevated risk of depression \cite{EberhardGran2002}, but are no more likely than non-pregnant women to receive diagnosis or treatment if they experience major depression, with over a third going undiagnosed despite much more interaction with the healthcare system during pregnancy \cite{ko2012depression}. Intimate partner violence is likely under-reported across all of the affected population \cite{emery2010examining}. Traditional detection approaches require regular administration of burdensome diagnostics, often followed by interviewing methods that make patients uncomfortable \cite{howard2014non}. The added social expectations of pregnancy and new motherhood as a time of joy and celebration may compound the difficulty of psychosocial risk disclosure. 

The current work extends prior findings that written language can provide insight into an individual's cognitive and affective mental states. In a test to discern genuine from simulated suicide notes, Pestian \textit{et al.} exceed human performance with decision trees using emotion words, part of speech tags, writing complexity, and readability metrics \cite{pestian2010suicide}. Other work has found linguistic elements in social media data that point to emotional distress, particularly depression and cases of heightened suicide risk. Paul and Dredze \cite{Paul_Dredze_2011} and Coppersmith \textit{et al.} \cite{Coppersmith2014} both show success in predicting Twitter reports of self-identified mental health issues like depression, with the former work using health-focused topic models and the latter word-based sentiment features along with unigram and 5-gram language models. Some recent work predicting suicide risk \cite{zirikly-etal-2019-clpsych} uses neural network frameworks, with models building on features generated from large datasets (e.g. word embeddings) \cite{mohammadi-etal-2019-clac-clpsych}, from human knowledge (e.g. sentiment and thematic features) \cite{allen-etal-2019-convsent}, or both \cite{matero-etal-2019-suicide}. A comparison of numerous models and feature sets by Tadesse \textit{et al.} demonstrates greater accuracy of mental health risk prediction when combining multiple types of features \cite{tadesse2019detection}.

\begin{table}[htb]
    \caption{Open-ended \& multiple choice survey questions} \label{tab:surveyq}
	\centering
	\begin{tabular}{l}
		\toprule
		\textbf{Free-text response questions}\\
		\textit{How would you describe your overall mood in the past}\\
		\kern10pt \textit{ 24 hours? What had the biggest impact on your mood,} \\
        \kern10pt \textit{and why?}\\
		\textit{In looking back at the past 24 hours, what events or }\\ 
		\kern10pt \textit{interactions stand out? How did they make you feel?} \\
		\textit{What activity or event did you most enjoy in the past }\\
		\kern10pt \textit{24 hours? What did you enjoy the least?  Why?} \\
		\textit{How have you been feeling about your pregnancy in} \\
		\kern10pt \textit{the last 24 hours? [or, if postpartum] How have you}\\
		\kern10pt \textit{been feeling about being a mother in the last 24 hours?}\\
		\midrule
		\textbf{Multiple choice questions}\\
		\textit{How would you describe your mood over the past 24} \\
		\kern10pt \textit{hours?} Very poor - Poor - Neutral - Good - Very good \\
		 \textit{In the past 24 hours, how much relationship conflict have} \\
		 \kern10pt \textit{you and your partner experienced?} \\
		 \kern10pt No conflict at all - Some conflict - Moderate conflict - \\
		 \kern10pt Substantial conflict - Extreme conflict \\
		 \textit{How would you describe your energy level over the}\\
		 \kern10pt \textit{past 24 hours?} Very low - Low - About normal \\
		 \kern10pt  - High - Very high \\
		 \textit{About how many hours did you sleep last night?} 0 - 1 - \\
		  \kern10pt 2 - 3 - 4 - 5 - 6 - 7 - 8 - 9 - 10 - 11 - 12 or more \\
		 \textit{How would you rate your sleep quality overall last night?}\\
		  \kern10pt Very bad - Bad - Good - Very good\\
		\bottomrule
	\end{tabular}
\end{table}

\begin{table}
    \caption{Psychological aggression screening questions} \label{tab:webq}
	\centering
	\begin{tabular}{l}
		\toprule
		\textbf{Psychological aggression questions}\\
		\textit{How much do you agree with the following statements?}\\
		\textit{I try not to `rock the boat' because I am afraid of what }\\
		\kern20pt \textit{my partner might do.} \\ 
		\kern10pt Strongly disagree - disagree - agree - strongly agree \\
        \textit{I feel owned or controlled by my partner.} \\ 
		\kern10pt Strongly disagree - disagree - agree - strongly agree \\
		\textit{My partner can scare me without laying a hand on me.} \\ 
		\kern10pt Strongly disagree - disagree - agree - strongly agree \\
		\bottomrule
	\end{tabular}
\end{table}

Most work in this space uses human classification of text as ground truth rather than psychometrically validated measures or clinical diagnoses. There are a few exceptions. Cook \textit{et al.} predict quantified suicidal ideation and psychiatric symptoms in a high-risk population based on \textit{n}-grams from a single open-ended question \cite{Cook2016}. Using sentiment and behavioral attributes, De Choudhury \textit{et al.} predict depression measure results using sentiment or grammatical features on a pre-onset social media dataset focused on the perinatal period \cite{DeChoudhury2014}. The latter study finds good predictive accuracy using hundreds of public posts per user, although depressed respondents often stated they tried not to disclose their difficulties online. 

Our approach assesses the value of using natural language processing techniques for indirect psychosocial risk detection based on a single structured interaction. To do this, we focused on what can be gleaned from private thoughts collected on a single day. 
These thoughts may include sensitive information that people would not share with friends and family, but also provide a smaller volume of text data per person examined than do most social media studies. 
We tested the usefulness of individual and combined language features, including affect, topic models, and thematic and grammatical terms to determine whether any or all of these features would be predictive on specifically elicited data with attached ground truth. As a basis for comparison about how much could be gleaned from a single interaction, we also tested categorical responses to closed-form questions as alternative predictors for psychosocial risk. 

\section{Data Collection}

Pregnant ($n=178$) and postpartum ($n=131$) participants were recruited through fliers at outpatient clinics serving pregnant women, as well as social media posts on groups for new and expecting mothers. Respondents were eligible to participate if they were 18-45 years of age and currently pregnant or had given birth within the past 12 weeks. Every respondent who was eligible to participate completed the survey. Participants received a \$25 Amazon gift card. Demographic data was collected at the end of each survey. 

\subsection{Text elicitation}

To improve the odds that respondents would complete the survey undisturbed, they were encouraged to do it in a quiet and private place. The survey began with four open-ended questions about the respondent's day, mood, recent experiences, and feelings about pregnancy or motherhood. Questions were framed to focus writers' attention on their emotional responses to experiences they described. To promote longer responses, they were asked to think of these questions as journal prompts and to focus most of their effort for the survey on this section. The writing prompts and other questions for independent variables can be found on the left side of Table \ref{tab:surveyq}. 

\subsection{Multiple-choice questions}
Following the open-ended questions, respondents answered multiple choice questions about mood, conflict in their relationship, energy level, and sleep quality. 

\begin{figure*}
    \centering
    \includegraphics[scale=.525, trim={0 3cm 0 0}]{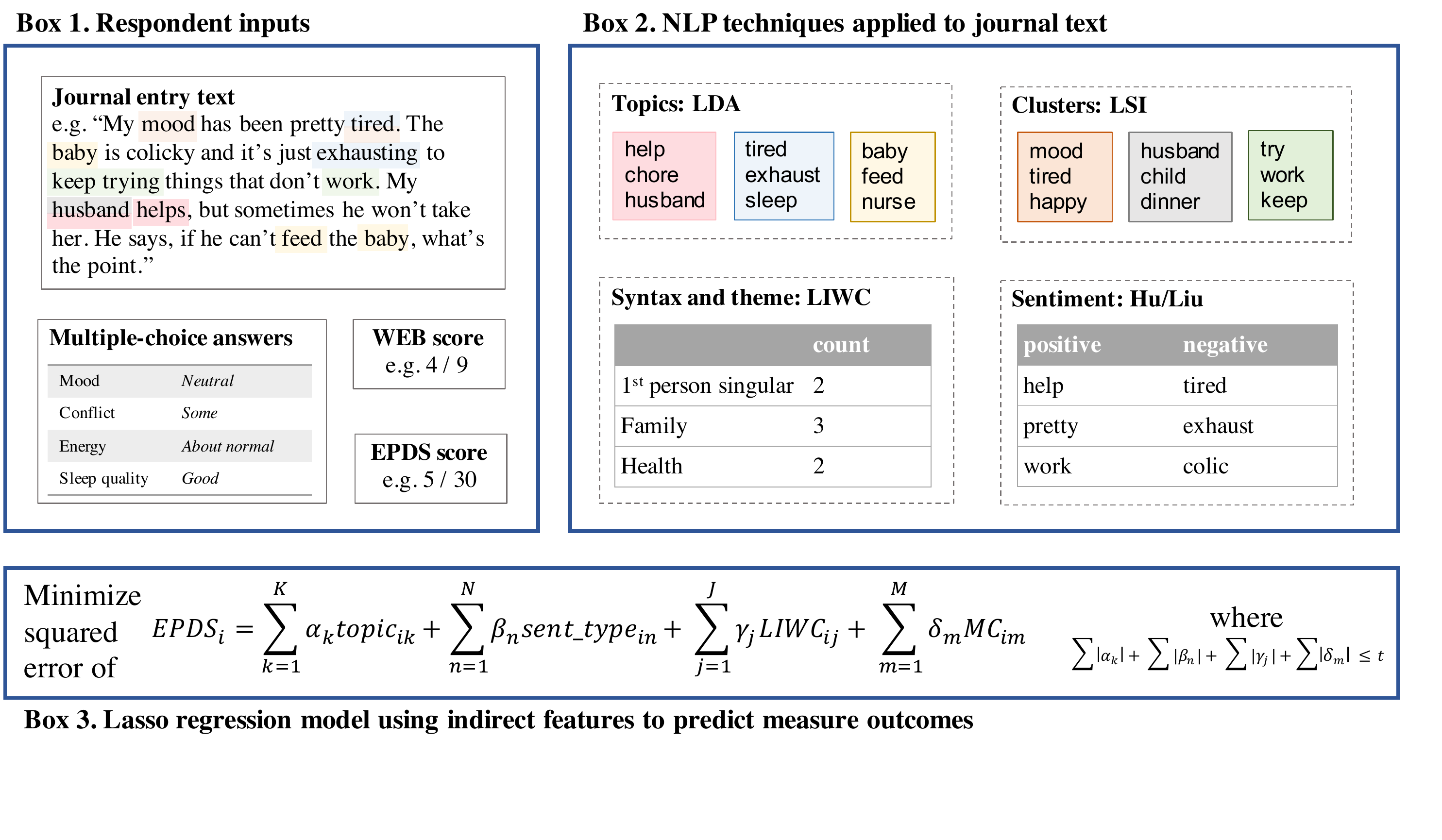}
    \caption{Create features and predictions from raw input.}
    \label{fig:process_flow}
\end{figure*}

\subsection{Risk measures} 

All respondents completed the 10-item Edinburgh Postnatal Depression Scale (EPDS) \cite{cox1987detection}, including an item about thoughts of self-harm to measure suicidal ideation. The EPDS cutoff was 13 or more, shown in the original study to indicate depression in postnatal women \cite{cox1987detection}. 

All respondents completed three items adapted from the Women's Experience with Battering (WEB) scale \cite{Smith1999} to measure psychological manipulation and aggression by a romantic partner on a four-point Likert scale, where agreement with at least one out of three items was labeled as psychological aggression, a type of IPV \cite{breiding2015intimate}. 

\subsection{Pre-registration and ethical considerations}

This work was pre-registered on the Open Science Framework at
\href{https://osf.io/uynsf/?view_only=45fcec7f32d04568b26a2ac411401336}{https://osf.io/uynsf/?view_only=45fcec7f32d045 68b26a2ac411401336}.
We deviated from the pre-registered protocol by adding LIWC features after pre-registration. The experimental design was approved by Carnegie Mellon University?s IRB (STUDY2017\_00000148) and all participants provided informed consent. All participants were offered resources for depression and partner violence after completing that section of the survey and were informed that these were offered regardless of their answers to preceding questions.

\section{Analytic Approach} 

The approach examined which aspects of natural language correlate with established psychometric measures of risk, and to what degree. We explored the separate and joint ability of indirect closed-form multiple choice questions and language features drawn from open-ended text responses to identify depression and intimate partner violence during the peripartum period. 

Figure \ref{fig:process_flow} summarizes the process and means of data collection and natural language feature analysis. Box 1 shows a simulated quote similar to a postpartum woman's open-ended text response to the question ``What events have most impacted your mood in the past 24 hours?'' above risk measure results. Box 2 notes the three natural language processing (NLP) techniques used, with latent Dirichlet allocation (LDA) and latent semantic indexing (LSI) each capturing topics of the journal entry and sentiment analysis estimating the positive and negative connotations of words used. Each natural language algorithm output a set of features that were scaled, then entered into a Lasso regularized regression model, shown in Box 3.

\subsection{Feature extraction}
Language features may reveal topics, affect, syntactic content, and thematic content. Topic modeling was selected in order to extract patterns in word use, which have been shown to be informative in classifying mental health content \cite{tadesse2019detection}. 
The other feature sets were generated as direct and indirect distillations of human knowledge about emotional and psychological connotations of particular words \cite{hu2004mining, baccianella2010sentiwordnet, tausczik2010psychological}. By incorporating these features we partially bypassed a common dilemma in classifying mental health text: datasets are too small to allow sufficient training of a semantically uninformed model.

\textit{Topic modeling}. Topic models were built on all of the text entries, including withheld data, using Latent Dirichlet Allocation (LDA) and Latent Semantic Indexing (LSI). 

\newcommand{\tabA} {
\scriptsize
    \begin{tabular}{ l | l | l } 
         \toprule
         & Pregnant & Postpartum \\
         & $n=178$ & $n=131$ \\
         \midrule
         \textbf{Age} \\
         18-24 years & 8\% & 22\% \\
         25-31 & 55\% & 42\% \\
         32-38 & 33\% & 34\% \\
         39-45 & 4\% & 2\% \\
         \midrule
         \textbf{Status} \\
         Not in & 0\% & 2\% \\
         \kern5pt relationship & & \\
         Never married & 12\% & 14\% \\
         Married & 85\% & 77\% \\
         Separated & 1\% & 2\% \\
         Divorced & 2\% & 7\% \\
         \midrule
         \textbf{Race} \\
         White & 88\% & 91\% \\
         Black & 3\% & 2\% \\
         Asian / Pac. Isl. & 2\% & 1\% \\
         Latina & 2\% & 2\% \\
         Multiracial & 4\% & 4\% \\
         Other & 1\% & 0\% \\
         \midrule
         \multicolumn{2}{l}{\textbf{Household income}} \\
         Under \$15k & 2\% & 11\% \\
         \$15-25k & 7\% & 10\% \\
         \$25-35k & 5\% & 11\% \\
         \$35-50k & 16\% & 15\% \\
         \$50-70k & 25\% & 20\% \\
         \$70-100k & 16\% & 15\% \\
         \$100k+ & 29\% & 17\% \\
         \bottomrule
	\end{tabular}
}

\newcommand{\tabB} {
\scriptsize
    \begin{tabular}{lccc} 
         \toprule
         \textit{Variable measured} & \textit{Range} & \textit{Mean} & \textit{SD}\\
         \midrule
         Depression (EPDS) & 0-30 & 7.1 & 4.8\\
         \kern10pt Suicidal ideation (EPDS item 10) & 0-3 & 0.22 & 0.56  \\
         Psychological aggression (WEB) & 0-9 & 0.66 & 1.3\\
         \textit{Multiple choice questions} \\
         Mood & 0-4 & 2.6 & 0.94 \\ 
         Partner conflict & 0-4 & 0.50 & 0.79 \\
         Energy & 0-4 & 1.5 & 0.82 \\
         Hours of sleep & 0-12 & 6.2 & 1.8 \\
         Sleep quality & 0-3 & 1.6 & 0.73 \\
         \bottomrule
         \multicolumn{4}{l}{EPDS cutoffs are $>9/30$ for mild depression; $>12/30$} \\
         \multicolumn{4}{l}{for moderate to severe depression  \cite{Murray1990}.} \\
    \end{tabular}
}

\newcommand{\tabC} {
\scriptsize
    \begin{tabular}{lll | ll} 
        \toprule
         & \multicolumn{2}{c}{Likely} & \multicolumn{2}{c}{Psychological} \\
         & \multicolumn{2}{c}{depression} & \multicolumn{2}{c}{aggression}  \\
         & \multicolumn{2}{l}{OR}{CI} & \multicolumn{2}{l}{OR}{CI} \\
         \midrule
         First pregnancy & 0.9 & 0.4-2.3 & 0.7 & 0.3-1.9 \\
         \hline
         Married & 1.1 & 0.4-3.4 & \textbf{0.3} & 0.1-0.8  \\\hline
         Urban area & 1.3 & 0.6-3.0 & 0.8 & 0.3-2.1 \\
         Suburbs & 0.7 & 0.3-1.6 & 0.8 & 0.3-1.8 \\
         Rural area & 1.1 & 0.4-2.6 & 1.6 & 0.6-3.7  \\\hline
         White & 0.8 & 0.3-2.7 & 0.9 & 0.3-4.1  \\
         Nonwhite & 1.3 & 0.4-3.7 & 1.1 & 0.2-3.3  \\\hline
         Household income $<$ \$50k & \textbf{2.3} & 1.1-5.0 & 2.1 & 1.0-4.9 \\
         \hline
         Bachelor's or higher & 0.5 & 0.2-1.1 & 0.4 & 0.2-1.0  \\
         \bottomrule
         \multicolumn{5}{l}{Confidence intervals shown are all 95\% intervals.}\\
    \end{tabular}
}

\begin{table*}[htb]
    \centering
    \caption{Descriptive statistics of survey data from 309 peripartum female U.S. residents.}
    \sbox{\measurebox}{
      \begin{minipage}[b]{.37\textwidth}
      \subfloat
        [Respondent demographics \label{tab:figA}]
        {\tabA}
      \end{minipage}}
    \usebox{\measurebox}\qquad
    \begin{minipage}[b][\ht\measurebox][s]{.50\textwidth}
    \subfloat
      [Distribution of scores from multiple choice questions and psychometric measures\label{tab:figB}]
      {\tabB} \\
    \subfloat
      [Demographics associated with depression and psychological aggression\label{tab:figC}]
      {\tabC}
    \end{minipage}
    \label{tab:descriptive}
\end{table*}

\textit{Latent Dirichlet Allocation}. LDA iterates over a library of documents to model the distribution of words within unseen underlying topics \cite{Blei_Ng_Jordan_2003}. Its outputs are sets of grouped words representing recurring topics of discussion in the documents. Each word in a topic has a score denoting its representativeness for that topic. LDA modeling used the R \cite{rlang} package \texttt{topicmodels} by Gr\"un and Hornik \cite{tmodels2011}, which wraps the C++ Gibbs sampler by Phan et al. \cite{phan2007gibbslda++}. 

A key parameter in LDA is $k$, the number of topics represented in the library of documents. To select an appropriate value, we used a measure of pairwise \emph{coherence} for each pair of words in the top 5 conditional likelihoods given a topic. Provided by Jones in \texttt{textmineR} \cite{textmineR} and independently described by Rosner \textit{et al.} \cite{rosner2014evaluating}, this probabilistic measure determines the increase in likelihood for each term when conditioned on other terms in the topic. Newman \textit{et al.}'s 2010 comparison showed that this type of mutual information measure best matched human annotator judgments about topic usefulness \cite{newman2010automatic}. In contrast, the more commonly used \textit{perplexity}, which estimates the likelihood of seeing withheld documents if the underlying topics match those predicted, tests the utility of the topics in document generation \cite{wallach2009evaluation} but yields topics that are less intelligible to human readers. By instead using word-specific mutual information of top terms in topics, coherence tests internal cohesion of each topic. Coherence was measured with $k=2$ to $k=100$ topics, and the final topics were generated with the $k$ maximizing average coherence. Given these topics, conditional likelihood (gamma) scores assessed every document's match with each topic. 

\textit{Latent Semantic Indexing}. LSI uses singular value decomposition on a matrix of raw word counts per document. Outputs are a matrix mapping concepts to words, another mapping documents to concepts, and a diagonal matrix indicating concept frequency. The document-concept matrix provided concept features for outcome prediction on each response.

\textit{Sentiment analysis}. The affective valence expressed in each entry was measured using two dictionaries. For the SentiWordNet word affect corpus \cite{baccianella2010sentiwordnet}, fractional positive and negative affect for each word (averaged across all homonyms provided) were averaged into statement scores for positive and negative affect. Opinion Lexicon scores were the number of statement words in the positive and negative lists divided by the total number of non-stopwords in each statement. 

\textit{LIWC features}. The Linguistic Inquiry and Word Count dictionary (LIWC) includes affect features, grammatical structures like pronouns and common verbs, and themes including perception, health, and money \cite{tausczik2010psychological}. Combining these counts with other sentiment and topic information added more features that incorporate human intelligence and are validated on large datasets, resulting in a feature set that more broadly reflected what a reader might take away from the writing. All 68 available LIWC features were used.

\textit{Categorical questions}. Responses to the five multiple-choice questions were treated as linear variables quantifying the respondent's subjective recent experience.

\subsection{Prediction with lasso regression}
Lasso regression allowed us to study the utility of feature sets and of individual features within sets. To test the performance of distinct types of features, seven aggregated sets were examined: sentiment scores (four features), LIWC counts (68 features), LDA results ($k=41$ features), LSI results (309 features), all-NLP (sentiment, LIWC, LDA, and LSI combined), multiple-choice questions (five features), and all-features. A consistent 20\% test set was split out in advance. Binary answers were generated with a cutoff of EPDS $\geq 13$ and any `Agree' answer for WEB \cite{cox1987detection, Chang2011, Murray1990}. For each set of features, \texttt{glmnet} \cite{glmnet} was used to perform lasso regression 100 times in a five-fold cross-validation on the training set. Loops assigned each feature a coefficient, often zero, and calculated an area under the ROC curve (AUC) using \texttt{ROCR} \cite{rocr}. To select the most useful individual features, coefficients were averaged across all loops to predict risk on withheld testing data \cite{gentzkow2019text}. Predictive performance was calculated using AUC. 

\section{Results}

Respondent demographics are shown in Table \ref{tab:figA}. 
Postpartum respondents differed from pregnant respondents on several dimensions, being generally younger, lower-income, and less educated.

\begin{table*}[htb]
\small
    \centering
    \caption{Averaged LASSO model performance.}
    \begin{tabular}{l|l|rr|rr}
        \toprule
         Risk & Feature class & \multicolumn{2}{c|}{5-fold cross-validation} & \multicolumn{2}{c}{Test set}\\
         & & \multicolumn{2}{c|}{on tuning set} & \multicolumn{2}{c}{} \\
         & & \multicolumn{2}{c|}{(100 replicates)} & \multicolumn{2}{c}{} \\
         & & \multicolumn{1}{c}{$R^2$} & \multicolumn{1}{c |}{AUC} & $R^2$ & AUC \\
         \midrule
         EPDS score & Pos, neg affect & .12 (.09) & .63 (.10) & .09 & .72 \\
         \textit{(depression)} & LIWC & .06 (.06) & .61 (.09) & .07 & .75 \\
         & LDA topics & -.01 (.04) & .51 (.10) & .02 & .61 \\
         & LSI topics & .00 (.02) & .49 (.06) & .00 & .63 \\
         & All NLP & .08 (.04) & .58 (.10) & .07 & .71 \\
         & Multiple choice & \textbf{.26} (.11) & \textbf{.70} (.11) & \textbf{.32} & \textbf{.79} \\
         & All features & .21 (.13) & .66 (.13) & .26 & \textbf{.79} \\ 
         \midrule
         WEB score & Pos, neg affect & .00 (.09) & .63 (.15) & -.01 & .45 \\
         \textit{(psych. aggression)} & LIWC & -.02 (.05) & .55 (.15) & -.01 & .45 \\
         & LDA topics & .00 (.01) & .46 (.09) & .00 & .67 \\
         & LSI topics & -.01 (.03) & .49 (.11) & .00 & .49 \\
         & All NLP & -.01 (.02) & .50 (.12) & .00 & .56 \\
         & Multiple choice & \textbf{.08} (.09) & \textbf{.72} (.14) & \textbf{.15} & \textbf{.80} \\
         & All features & .04 (.06) & \textbf{.72} (.15) & .08 & .75 \\ 
         \midrule
         \multicolumn{6}{l}{Note: Testing was performed on withheld 20\% of data, using coefficients}\\
         \multicolumn{6}{l}{averaged from all replicate regressions. For EPDS, indicator variable}\\
         \multicolumn{6}{l}{threshold for AUC was a score of 13 or higher. For WEB, it was} \\ 
         \multicolumn{6}{l}{a response of ``agree'' or ``strongly agree'' on at least one question.}\\
         \multicolumn{6}{l}{Best-performing results are bolded.}\\
         \bottomrule
    \end{tabular}
    \label{tab:lasso}
\end{table*}

Table~\ref{tab:lasso} shows the performance of feature groups used separately or together to predict the two psychosocial risks measured. 

\subsection{Topics}

\begin{figure*}
     \centering \small
     \begin{tabular}{l | l | l | l | l} 
     \toprule 
     Topic 4 & Topic 7 & Topic 9 & Topic 12 & Topic 13\\
     \textit{Pregnancy} & \textit{Positive} & \textit{Process focus} & \textit{Family} & \textit{Anticipating}\\
     
     \kern5pt \textit{anxiety} & \kern5pt \textit{duties} & & \kern5pt \textit{relationships} & \kern5pt \textit{delivery}  \\
     \midrule
     anxious & toddler & get & kid & excite\\
     week & house & last & older & first\\
     day & nice & pregnancy & interact & pretty\\
     house & walk & ready & mom & due\\
     hope & around & eat & newborn & nervous\\
     birth & went & week & sit & normal \\
     appoint & outside & got & frustrate & sometimes \\
     doctor & next & stop & asleep & ultrasound \\
     & & & & \\
     \toprule
     Topic 17 & Topic 26 & Topic 27 & Topic 33 & Topic 40 \\
     \textit{Empowered} & \textit{Everything in} & \textit{Feeling busy} & \textit{Nurturing} & \textit{Lack of} \\
     \kern5pt \textit{pregnancy} & \kern5pt \textit{perspective} & & & \kern5pt \textit{support} \\
     \midrule
     baby & happy & time & take & husband \\
     day & make & spend & care & sick \\
     able & see & just & happy & help \\
     clean & two & overwhelmed & nurse & morning \\
     think & different & change & try & anything \\
     say & dinner & say & motherhood & really \\
     pregnant & healthy & happy & every & come \\
     smile & now & start & feed & sad \\
     \bottomrule
     \end{tabular}
     \caption{Top words from selected topic model output from LDA on pregnant and postpartum respondents' statements. $k=41$ optimized coherence. Stemmed words were manually expanded here for clarity (e.g. `morn' $\rightarrow$ `morning'). Topic headings as interpreted by authors.} \label{fig:tmwords}
\end{figure*}

Coherence in LDA models peaked at $k=41$ topics, some of which are shown in Figure \ref{fig:tmwords}. Regularized regression with LDA topics indicated that topics 9, 26, and 7 were most useful for predicting depression, having average coefficients of 19.50, -11.37, and -10.55, respectively. Six LDA topics were chosen by lasso in more than half of its loops, while only one LSI topic was used more than one fifth of the time. For predicting psychological aggression, eight LDA topics had averaged coefficients above 0.01. Topic 40 was most informative, with an average coefficient of 0.23 and being selected in 34 out of 100 model runs. The next most frequent LDA topic was topic 17, selected only 4 times out of 100.

Regression with LDA topics performed above chance for both measures, while using LSI topics surpassed chance only when regressing on depression. 

\subsection{Sentiment}
As shown in Table~\ref{tab:lasso}, a model built with sentiment features performed better than LDA features on EPDS, and worse than LDA features on WEB. Using more words on the Opinion Lexicon (OL) positive word list \cite{hu2004mining} was associated with a lower EPDS score, with a $-1.12$ mean regression coefficient for the scaled ratio of positive words, compared to $-.028$ for scaled positive word scores based on SentiWordNet (SWN) \cite{baccianella2010sentiwordnet}. The coefficient difference between the sentiment dictionaries was smaller for negative sentiment scores, where SWN's average coefficient for EPDS was .21, compared to .87 for the negative OL feature. Lasso also selected SentiWordNet features less frequently: when only the four sentiment features were considered in predicting depression, negative affect and positive affect from SWN were used 79/100 times and 24/100 times, respectively. Both OL features were used every time. 

When predicting psychological aggression, lasso's preference for OL features was still present but less pronounced. Negative OL terms were associated with a higher chance of psychological aggression and positive terms implied a lower chance. The relationship was reversed for SWN features, where positive terms predicted higher risk and negative words suggested lower risk. 

Together, a regression with the four sentiment features showed better-than-chance predictive accuracy on EPDS, with an AUC of 0.63 on cross-validation and 0.72 on the test set. Combining sentiment and topic features did not meaningfully improve predictive ability. 

\subsection{Thematic and grammatical features}

LIWC features used alone performed as well as affect features alone when predicting depression. The features did not appear to capture meaningfully distinct information, showing similar performance when they were used independently or combined with the affect features. 

\begin{figure}
    \centering
    \begin{minipage}{0.5\textwidth}
        \centering
        \includegraphics[width=0.9\textwidth]{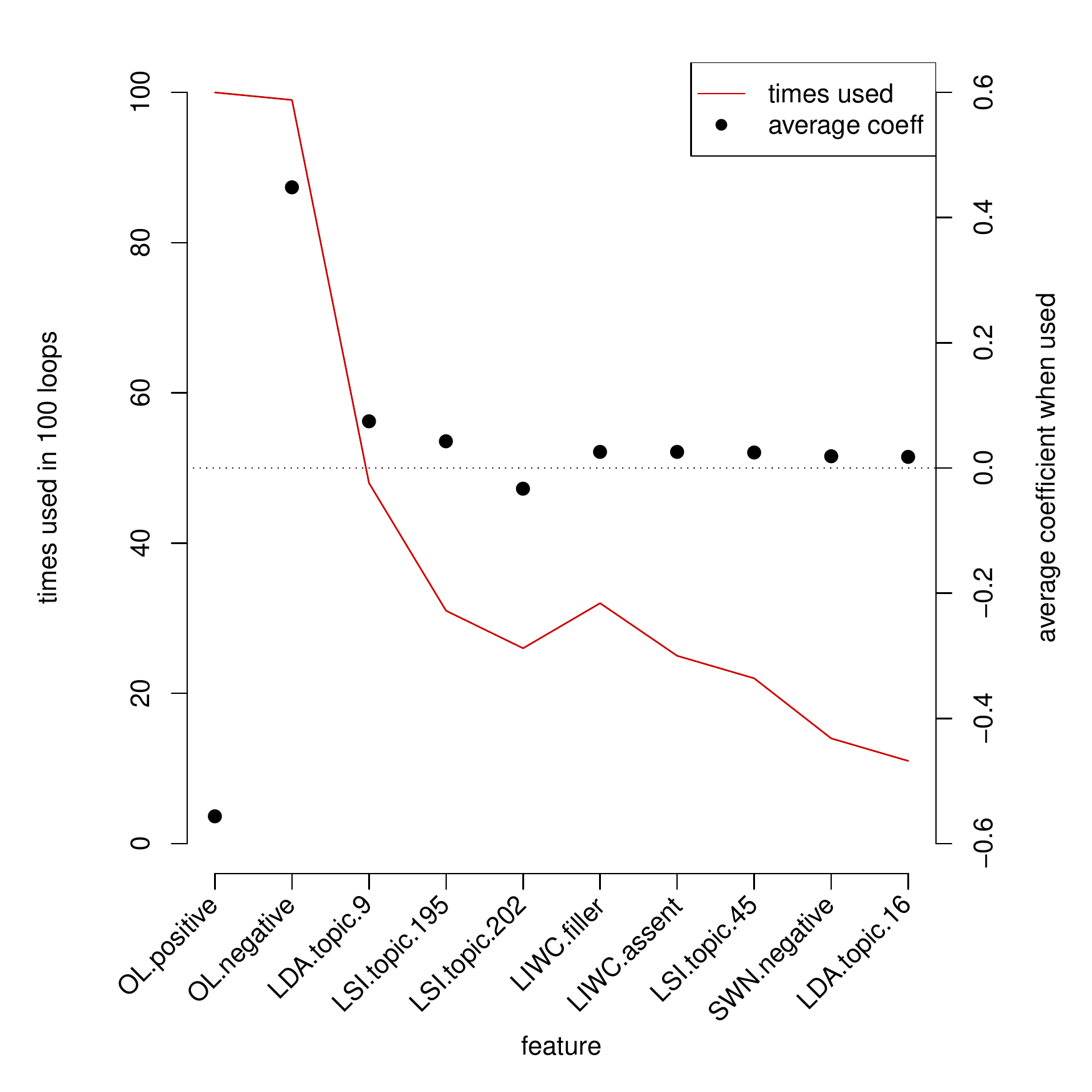} 
        \caption{Language features for depression prediction} \label{fig:coefs}
    \end{minipage}\hfill
    \begin{minipage}{0.5\textwidth}
        \centering
        \includegraphics[width=0.9\textwidth]{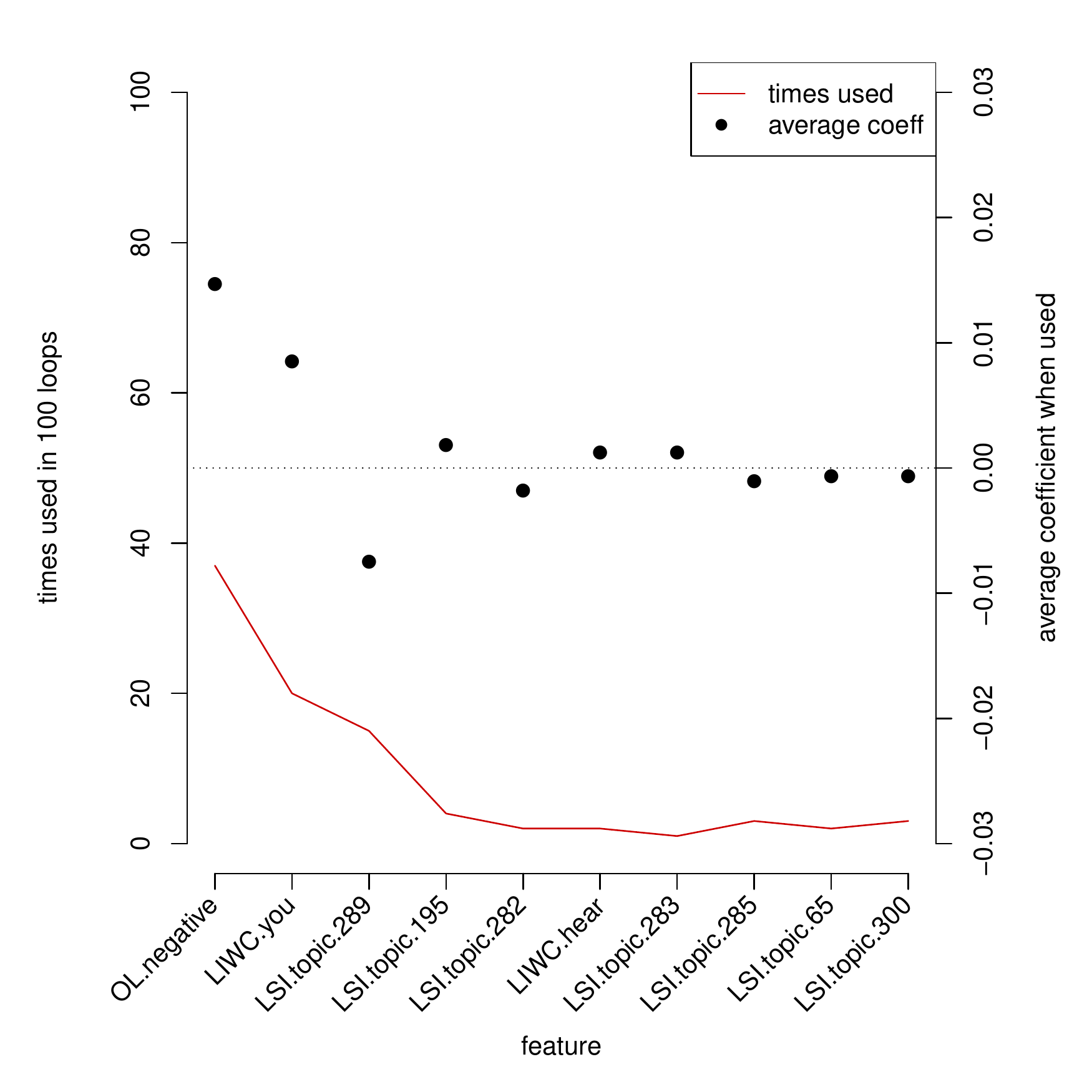} 
        \caption{Language features for psych. aggression prediction} \label{fig:coefs2}
    \end{minipage}
\end{figure}

A combination of all other NLP features---affect, LDA topics, and LSI topics---was evaluated in comparison to all of these features with LIWC added, and performance was nearly identical on both depression and partner violence. The specific features favored by lasso changed, with three LIWC features (`filler', `assent', `anger') appearing among the twenty most-selected in 100 model runs on depression. For intimate partner violence, the `you' feature (second-person pronoun use) appeared in a fifth of the model runs, making it the second most-used feature behind negative OL affect. The top features chosen for each measure are shown in Figures \ref{fig:coefs} and \ref{fig:coefs2}.

\subsection{Closed-form indirect measures}

The second-to-last row of Table~\ref{tab:lasso} shows that, compared with natural language features, predictions using multiple choice questions tended to be more strongly correlated with WEB than were natural language models. The same finding was true, though less exaggerated, for the EPDS; when predicting depression, individual feature correlations were
$R^2=0.23$ for mood, $R^2=0.11$ for partner conflict, $R^2=0.08$ for energy, $R^2=0.09$ for sleep quality, and $R^2=0.08$ for hours slept. Partner conflict was most highly correlated with the WEB ($R^2=0.14$), while mood, hours slept, and sleep quality each had an $R^2$ between 0.01 and 0.02. 

Coefficients for variables in the all-features lasso model of EPDS showed that the multiple choice questions were given less weight in that model than when considered alone: for instance, in this model the `conflict' variable had a coefficient of 0.83, while in the all-features model it was 0.35. The `mood' variable similarly dropped from a -1.8 coefficient to -1.4. A similar pattern holds for WEB and the conflict question; its coefficient is 0.33 in the multiple-choice-only model, and 0.17 in the all-features model, although the latter does slightly worse on the test set. 

\section{Discussion}

This work addresses the dilemma of identifying harmful but stigmatized risks in a perinatal population, where direct disclosure can be difficult but few alternatives exist for detection. 
We tested two techniques to elicit and analyze indirect information that may predict risks of depression and intimate partner violence. 
Building on Coppersmith \textit{et al.} \cite{Coppersmith2014}, Cook \textit{et al.} \cite{Cook2016}, and De Choudhury \textit{et al.} \cite{DeChoudhury2014}, among many others, we extracted natural language features (topic, sentiment, and themes) from a single journal-style text sample to predict psychosocial risks. Overall, lasso regression on sentiment and thematic features of a woman's language allowed prediction of depression without directly asking about her depressed feelings. Intimate partner violence was more difficult to predict from natural language alone, but could be detected (AUC=0.80) from closed-form indirect measures like relationship conflict. In the next three sections, we discuss the findings with respect to context of other screenings, the utility of closed-form questions, and using sentiment to predict depression. 

\subsection{Performance in context}

The features extracted from closed-form and open-ended questions provided an indirect means of screening for psychosocial risks, expanding detection to contexts where the direct questioning approach of existing methods may fail. Both the closed-form and open-ended questions could be considered less invasive because they do not directly ask about depression and intimate partner violence, yet their predictive accuracy is similar to other direct questioning methods (e.g., PHQ2), although comparisons to prior work are problematic because studies vary in the criteria used to determine depression and psychological aggression. Our approach detected depression as measured by the Edinburgh Postnatal Depression Scale with an AUC of 0.79, showing similar accuracy as PHQ-2 (the ``Whooley questions"), which have AUCs of 0.76-0.79 on the Structured Clinical Interview DSM-IV-TR (assuming Gaussian ROC curves) \cite{howard2018accuracy}. 

\subsection{Superior performance by closed-form questions}

Responses to indirect closed-form questions were more predictive than language features derived from open-ended text for both risk types, with a larger gap between feature types for aggression prediction. Allowing use of the five multiple-choice answers led to an AUC around 0.8 for both outcomes, while allowing all NLP features yielded a 0.71 AUC for predicting EPDS and 0.56 for predicting WEB---the latter being not far above chance. Test set error was slightly lower than cross-validation error on the training set with most feature sets, likely because testing models were averaged across all training runs, while lasso minimized the variables used during training. One likely reason for this gap between closed-form and open-ended features is the small training set: it was easier for lasso to fit a good model on the smaller, closed-form feature set without overfitting, which would be penalized during cross-validation.

A likely key factor in the success of the closed-form questions was their selection based on previous domain knowledge about the two risks we explored: low mood and energy are frequent symptoms of depression \cite{dsm-v}. Similarly, `relationship conflict' is a documented precursor to IPV \cite{marshall2011enduring}, and the question may have yielded honest responses because answers do not directly ascribe blame. Prior work has found that more direct questioning can lead to evasive responses \cite{alvarez2018responding}. In contrast, while our methods allowed the \textit{kinds} of natural language features to span a wide range of affect and themes, the specific features were not pre-selected for relevance to depression and psychological aggression. Instead, topics were generated based on the few hundred documents, and sentiment and thematic features came from multi-purpose dictionaries. With many irrelevant features and few true positives, it should be expected that the lasso would struggle to separate noise from signal in the correlations between features and risk. Additionally, interactions between variables were not considered in this model, so more complex underlying themes that might reflect specific relevant thought patterns (e.g. the co-occurrence of negative affect and LIWC's `money' feature) would not be captured by the model. There is no way to counteract the issues of noise or missing interactions in this model without significantly expanding the dataset size. Future work might carefully select specific language features to test, or else must collect datasets an order of magnitude larger to allow application of more powerful natural language processing tools like word embeddings \cite{mikolov2013distributed}, long short-term memory networks \cite{dai2015semi}, tree boosting \cite{chen2016xgboost}, or deep bidirectional transformers \cite{devlin2018bert}.

Overall, while such expansion could be informative, there were patterns in survey responses that enabled detection from closed-form questioning. In contrast, in many cases it would be difficult for any trained human reader to predict the risk from this dataset: many respondents simply did not mention their partners or relationships in their responses, including some of those who were experiencing psychological aggression. In principle, the patterns picked up in closed-form questions could also be detected in natural language, if open-ended questions were designed to elicit those same disclosures. Open-ended questions would need to be tailored to such issues, while respecting patients' reticence to answer direct questions---healthcare providers are also often unable to elicit disclosure, even when they suspect it \cite{alvarez2018responding}. Asking specifically about writers' (or patients') interactions with partners, friends, and family may yield enough discussion of those relationships for risk prediction using natural language processing, while remaining indirect enough to allow oblique reference to the issues. A comparably directed question for depression may prompt self-reflective language by specifically asking about longer-term mood patterns, enjoyment in normal activities, focus, and energy. 

\subsection{Sentiment and theme are informative for depression but not aggression}

The most informative language features for predicting depression were derived from sentiment and thematic dictionaries from prior work. Positive and negative affect features, particularly those from the Opinion Lexicon (OL), were consistently correlated with depression. LIWC features also did well on their own, achieving almost as high an AUC as the closed-form questions, although without the correspondingly strong $r^2$. In comparison, both topic models were significantly worse than the closed-form question results. 

One advantage of these natural language dictionaries is that they were developed from large datasets in prior work, unlike the topic models, which used only the data collected from journal entries in this study. All three dictionaries were built using thousands of documents and benefited from human judgment in seeding their categories.

In prior work, LIWC in particular has been effectively used in predicting mental health status from language \cite{Coppersmith2014, allen-etal-2019-convsent, tadesse2019detection, DeChoudhury2014, althoff2016large, nguyen2014affective}. 
Despite having significantly more features available, encompassing thematic concepts as well as affect, LIWC features only slightly improved performance over the basic OL and SentiWordNet (SWN) affect features. Either almost all meaningful information from LIWC was captured by the other affect features, or performance could not be improved upon given the dataset's size and relative precision of measurements. 
Note that a depression link did not appear for the LIWC first-person pronoun feature, while prior work has shown a correlation \cite{holtzman2017meta}. Because all respondents were asked to write about their daily experiences, the amount of first-person language likely varied less than it generally would across depressed and non-depressed respondents.

Within the affect features, there was more consistent and stronger use of the Opinion Lexicon features than the SentiWordNet features. The less frequent selection of SentiWordNet items may have resulted from its less empirically-grounded derivation, drawing on an existing resource, WordNet \cite{christiane1998wordnet}, which encodes relationships between terms, such as hypernymy (superset-of), synonyms, and antonymy. These relationships do not necessarily capture similarities in expressed affect. For example, synonyms can be used to connote different valences, where near-synonyms express subtle affective differences (e.g., \textit{parched} is more negative than \textit{dry}). In addition, words pick up affect connotations from their frequent use in particular contexts. For instance, while \textit{cockroach} is a hyponym of \textit{insect}, its connotation is much more negative both because of a common, specific dislike for cockroaches and because \textit{insect} is a more technical and less emotionally loaded term. Any of these variations in use could lead to significant differences in the affect associated with synonyms. The Opinion Lexicon also partially draws on WordNet relationships to bootstrap its polarity estimations. However, it focuses on 6,000 terms that are meaningful for predicting opinions on online product reviews. In contrast, SentiWordNet annotates all 200,000+ definitions in WordNet, likely following much longer chains of synonyms, antonyms, and hyponyms. Among words occurring across pairs of dictionaries, SWN disagrees with both OL and LIWC about the polarity for 25\% of words \cite{potts2011sentiment}. In contrast, OL and LIWC disagree on only 2\% of their overlapping terms, and both were more favored by the lasso model. 

Sentiment was the best natural language feature for predicting depression, but was not predictive of psychological aggression by an intimate partner. This may reflect a masking of negative affect, where people experiencing intimate partner violence sometimes dissociate and do not consciously experience their negative feelings \cite{siegel2012splitting}.
Alternatively, negative affect may be experienced internally but not communicated externally due to the anticipated negative reactions from others. For example, expressing negative affect might prompt anger from an abusive partner, making a woman experiencing IPV prefer not to ``rock the boat'' \cite{Smith1999}. Capturing this kind of risk from language may be more effective with a specific prompt about their interpersonal and romantic relationships. Still, explicit acknowledgment of intimate partner violence might suggest the need for a major life change (housing, finances, custody battles, all with the chance of escalated violence) \cite{meyer2012women}, making the expression of negative affect related to intimate partner violence too costly for some people to be willing to disclose it. 

One modeled topic shows the informative power of discussing other elements of relationships: LDA topic 40, which reflects a lack of spousal support, was linked with psychological aggression. The link suggests a hypothesis: either psychological aggression co-occurs with having an unsupportive partner, or disclosing this issue is more common among women experiencing abuse, perhaps as a less direct way to allude to problems in the relationship. The creators of the WEB scale note that violence in abusive relationships is usually relatively infrequent---but its use is held out as a spoken or unspoken threat to ensure compliance by the abused partner \cite{Smith1999}. People using psychological aggression as a tool for control may lash out relatively infrequently, but could use the implied threat to regularly avoid expected duties in their relationships. In such cases, a single day's experiences may not contain overt aggression, but could still reflect the resultant imbalance in workload between partners. If this thematic link holds true in clinical practice, it will be useful for practitioners to know that women experiencing psychological aggression may disclose a general lack of support from their partners. 

\subsection{Limitations}

The size of this dataset (248 instances for training, 61 for testing) makes it difficult to train a complex model. Relatively few respondents in the training set score positively for the risks in question: 19 (8\%) for psychological aggression, 36 (15\%) for depression. In this context, trends can be driven by a single outlier: for instance, topic 40's connection with psychological aggression comes largely from a single high scorer on both dimensions. Because there are few instances of depression and psychological aggression in the dataset, these results should be interpreted as plausible hypotheses for clinical work, rather than definitive correlations. Any model attempting to capture a meaningful part of the countless dimensions of linguistic variation would struggle with a dataset of this size and distribution. Pre-training on a related dataset could alleviate this difficulty, but few such datasets exist because of the sensitive nature of the data. The majority of work trying to predict mental health risk from language uses social media data. The context of the writing examined here intentionally varies from social media data: rather than sharing with the world (even anonymously), participants were told their data would be contained to a single research group. As a result, they may have been more forthcoming, writing with a different target audience in mind. Still, pre-training on related and somewhat more abundant social media data may improve prediction. Future work may also benefit from collecting longitudinal data, adding more data per respondent and letting providers examine patterns of distress over time.

In addition to the difficulty in extracting meaningful information that is present in language, it is plausible in this case that some journal responses did not contain enough information for any detection method to use. Some respondents did not mention their relationships, and without any direct mentions it may be that low-risk and high-risk responses simply cannot be differentiated. As noted above, the specificity of writing prompts must be considered along with the dataset size.

While this work uses validated measures of depression and psychological aggression, a gold standard would include in-person clinical assessment. One risk with validated self-report measures is noise; for instance, false negatives can occur in cases of IPV, where admitting there is a problem can be costly \cite{white2015cross}.
Still, clear communication and anonymized collection lower the barriers to disclosure as much as possible, and it is reasonable to assume that respondents were mostly honest. Many volunteered sensitive issues like mood swings, eating disorders, feelings of inadequacy, and sexual functioning; it is credible that they were similarly forthcoming when asked about specific issues. 

Respondent demographics were not representative of the target population: in particular, the snowball sampling method led to a disproportionate number of white respondents. With a model calibrated using this dataset, it is possible that detection will be less accurate on a different population. In this case the misalignment is troubling because, when depression appears among African-Americans, it is more likely to be debilitating than when it appears in white Americans \cite{williams2007prevalence}, and less likely to be adequately treated \cite{rockville1999mental}. Any tools proposed to supplement patient screening should be tested to ensure accuracy among those least likely to be helped by conventional methods. To do so, future work should expand demographic collection to ensure that more vulnerable populations are well-represented. 

\subsection{Clinical context}

Respondents in this survey volunteered to share their information, and it was not used in any context linked to their identities. In a medical realm, future work should even more carefully consider the informed consent procedures in asking sensitive questions. Respondents may write in a `journal' application assuming it will be kept private, while providers want to intervene in cases of likely risk. Further research on how to balance privacy and safety in this domain is needed.

An interesting application for future work is risk detection from dialogue. In such cases, the interlocutor may either divert from disclosure or encourage it. Learning which responses encourage elaboration will be as critical as detecting indirect disclosures when they occur. 

\section{Conclusions} 

Successful detection of risks from closed-form questions and short journal entries suggests that indirect, self-administered elicitation can enable meaningful predictions about psychosocial risks without requiring respondents to wait for a meeting with a practitioner or to explicitly disclose depressive symptoms or abuse. Non-burdensome methods like these could substantially improve detection in between appointments and for people who are reluctant to directly discuss stigmatized risks. 

\section*{Acknowledgments}

This work was partially supported by the National Science Foundation Graduate Research Fellowship Program under Grant No. DGE1745016. Any opinions, findings, and conclusions or recommendations expressed in this material are those of the author(s) and do not necessarily reflect the views of the National Science Foundation. T. Krishnamurti was funded by the Henry L. Hillman Foundation and NIH KL2 TR001856. A. Davis was funded by the Riksbankens Jubileumsfond program on Science and Proven Experience. 
Dr Krishnamurti and Dr Davis are co-founders of NAIMA Health LLC, which may use this research for pregnancy healthcare management. A patent for this work has been filed.

\bibliographystyle{IEEEtran}
{\small
\bibliography{references}}

\end{document}